\documentclass[conference]{IEEEtran}
\usepackage{times}
\usepackage{multicol}
\usepackage{multirow}
\usepackage[bookmarks=true]{hyperref}
\usepackage{amsmath}
\usepackage{amsthm}
\usepackage{amssymb}
\usepackage{graphicx}
\usepackage[dvipsnames,table,xcdraw]{xcolor}
\usepackage[inkscapelatex=false]{svg}
\usepackage[font=small,labelfont=bf]{caption}
\usepackage{subcaption}
\usepackage{graphicx}
\usepackage{soul}
\usepackage{cancel}
\usepackage{physics}
\usepackage[normalem]{ulem}
\usepackage{booktabs}
\usepackage{soul}
\usepackage[numbers]{natbib}
\usepackage{multicol}
\usepackage[bookmarks=true]{hyperref}
\usepackage{float}
\restylefloat{table}

\newcommand*{\draft}{} 
\ifdefined\draft
    \usepackage{comment}
    \usepackage[textsize=footnotesize]{todonotes}
    \DeclareRobustCommand{\harshil}[1]
    {{\todo[color=orange!40,inline]{HP: #1}}}
    
    \DeclareRobustCommand{\yifei}[1]
    {{\todo[color=green!40,inline]{YS: #1}}}  
    
    \DeclareRobustCommand{\pratik}[1]
    {{\todo[color=pink!40,inline]{Pratik: #1}}}  
    
    \DeclareRobustCommand{\junyao}[1]
    {{\todo[color=yellow!40,inline]{JS: #1}}} 

    \DeclareRobustCommand{\chuye}[1]
    {{\todo[color=blue!40,inline]{CZ: #1}}}

    \DeclareRobustCommand{\nadia}[1]
    {{\todo[color=pink!40,inline]{NF: #1}}}

    \usepackage[nopar=false]{lipsum}
    
    \DeclareRobustCommand{\fake}[1]{\color{gray!40!}{\lipsum[][#1]}\color{black}}
    
\else 
\usepackage{pbalance} 
\newcommand{\yifei}[1]{}
\newcommand{\harshil}[1]{}
\newcommand{\pratik}[1]{}
\newcommand{\nadia}[1]{}
\newcommand{\chuye}[1]{}
\newcommand{\junyao}[1]{}
\newcommand{\fake}[1]{}

\fi

\providecommand{\Rtri}{\ensuremath \mathbb{R}^3}

\providecommand{\Rsix}{\ensuremath \mathbb{R}^6}
\providecommand{\Rsixsq}{\ensuremath \mathbb{R}^{6\times6}}

\providecommand{\Stri}{\ensuremath \mathbb{S}^3}

\providecommand{\uj}{\boldsymbol{\tau}\jnt}

\providecommand{\state}{\mathbf{x}}

\providecommand{\xpos}{\mathbf{p}}
\providecommand{\jacob}{\mathbf{J}}

\providecommand{\quat}{\mathbf{q}}
\providecommand{\rotv}{\boldsymbol{\omega}}

\providecommand{\jnt}{_\theta}
\providecommand{\jntv}{\boldsymbol{\theta}}

\newcommand{\des}{^\text{d}}

\providecommand{\impD}{\mathbf{D}}

\providecommand{\pos}{p}
\providecommand{\rot}{o}
\providecommand{\dsg}{\mathbf{x}^*}
\providecommand{\dsgpos}{\xpos^*}
\providecommand{\dsgquat}{\quat^*}

\providecommand{\dsA}{\mathbf{A}}
\providecommand{\dsAp}{\mathbf{A}_\pos}
\providecommand{\dsAr}{\mathbf{A}_\rot}
\providecommand{\dsf}{\mathbf{f}}
\providecommand{\dsu}{\mathbf{u}_{\text{DS}}}

\providecommand{\tune}{{\boldsymbol{\eta}}}

\providecommand{\decay}{d}

\providecommand{\est}[1]{\ensuremath \widehat{#1}}
\providecommand{\twonorm}[1]{\ensuremath ||#1||_2}

\newcommand{\stkout}[1]{\ifmmode\text{\sout{\ensuremath{#1}}}\else\sout{#1}\fi}

\begin{document}

\title{Don't Yell at Your Robot: Physical Correction as the Collaborative Interface for Language Model Powered Robots }

\author{
    \IEEEauthorblockN{Chuye Zhang$^*$, Yifei Simon Shao$^*$, Harshil Parekh, Junyao Shi,\\ Pratik Chaudhari, Vijay Kumar, Nadia Figueroa}
    \IEEEauthorblockA{GRASP Laboratory, University of Pennsylvania\\
    Philadelphia, PA, 19104 USA\\
    \tt\small chuyez@sas.upenn.edu, \{yishao, hparekh, junys, pratikac, kumar, nadiafig\}@seas.upenn.edu}
    \thanks{$^*$These authors contributed equally.}
}

\maketitle

\begin{abstract}

 We present a novel approach for enhancing human-robot collaboration using physical interactions for real-time error correction of large language model (LLM) powered robots. Unlike other methods that rely on verbal or text commands, the robot leverages an LLM to proactively executes 6 DoF linear Dynamical System (DS) commands using a description of the scene in natural language.
During motion, a human can provide physical corrections, used to re-estimate the desired intention, also parameterized by linear DS. This corrected DS can be converted to natural language and used as part of the prompt to improve future LLM interactions.
We provide proof-of-concept result in a hybrid real+sim experiment, showcasing physical interaction as a new possibility for LLM powered human-robot interface.
\end{abstract}

\IEEEpeerreviewmaketitle

\section{Introduction}

A longstanding goal in robotics has been to develop autonomous robots capable of working by themselves. \cite{octo_2023}  Although this goal may be realized in the future, current robotic systems still lack human-level capabilities in critical areas such as real-world scene understanding, dexterous manipulation, and long-horizon planning. Nevertheless, robots can still significantly reduce human effort and enhance productivity by handling heavy objects and working alongside humans. Therefore, we focus on creating systems that effectively complement human abilities.

 \begin{figure}[htbp]
    \centering
    \includegraphics[width=0.98\linewidth,trim={0.5cm 0.1cm 0.3cm 0.3cm},clip]{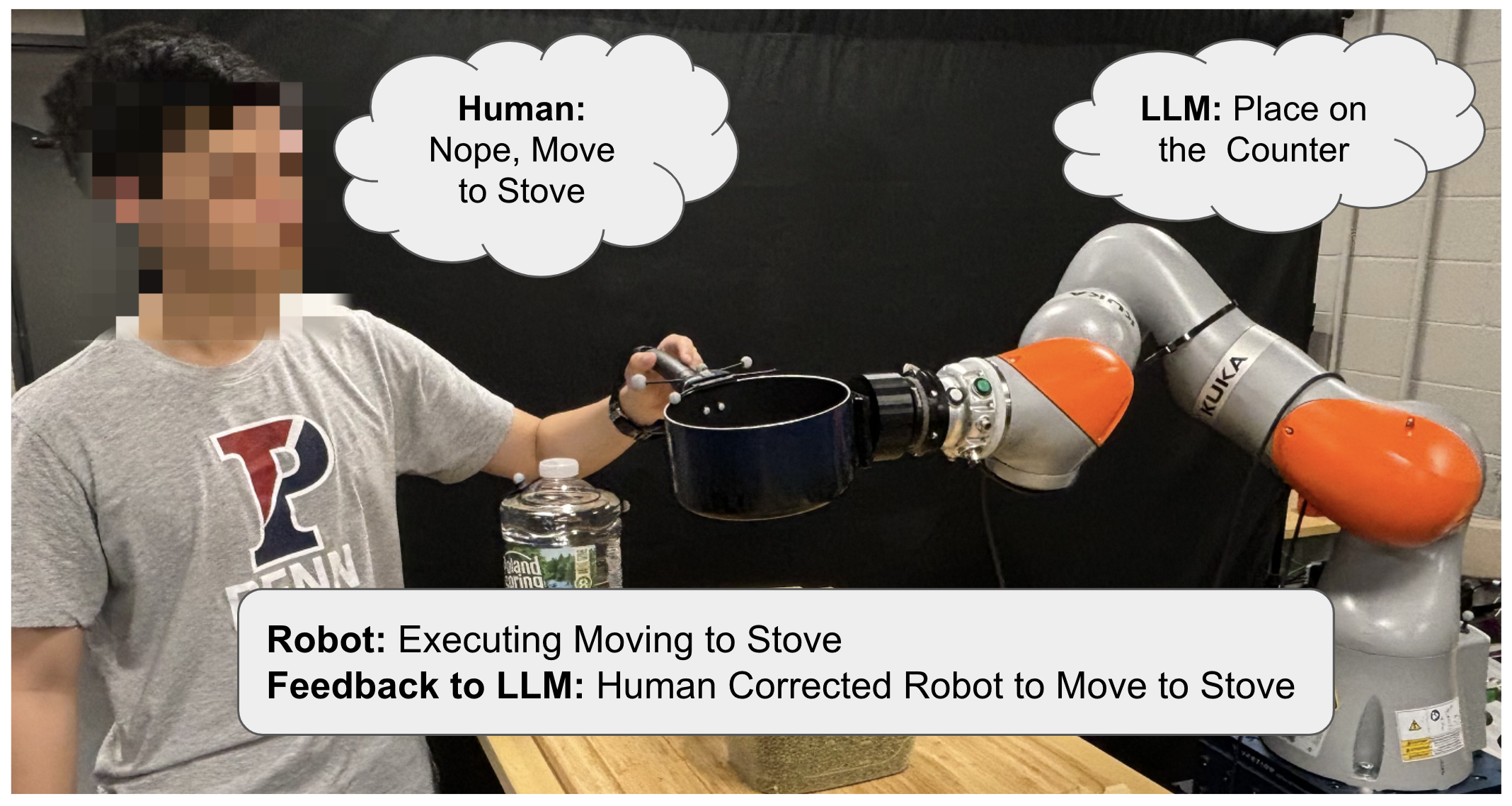}
    \caption{\small {The robot tries to proactively help by co-carrying a pot with the human. However, based on the current observation of the environment, the LLM outputs the incorrect action to move the pot towards the cutting board. With physical human correction, the robot adjusts the Dynamical System (DS) action parameters and aids human in carrying the pot towards the stove. A feedback of this correction is sent to the LLM for learning.}}
    \vspace{-17pt}
    \label{fig: robot_human_cooking}
\end{figure}
A recent approach is to use natural language for tasking and interacting with robots powered by large language models (LLMs)~\cite{ahn2022i, huang2022inner, Wu_2023, ma2024eureka, liang2023code, embodimentcollaboration2024open}. However, while LLMs excel in high-level commonsense reasoning, they struggle with consistently generating actions that are correct and aligned with human intentions.  Prior works have thus shown that using language as an interface for task specification can be delayed~\cite{lynch2022interactive}, underspecified~\cite{Cui_2023, shi2024yell}, and prone to errors~\cite{liu2023reflect, zha2024distilling, zheng2024evaluating, jiang2024transic}, especially performing long-horizon tasks in complex environments. 

However, using physical input as the collaborative interface has been under-explored.
One line of study uses physical corrections to learn human objectives through inverse reinforcement learning approaches  ~\cite{losey2022physical, pmlr-v48-finn16, bobu2021feature, li2021learning}. These methods are more usually aimed to obtaining autonomous behaviors aligned with human intention after correction, but usually require a learning phase and cannot be deployed in real-time.

We advocate for robots to assist humans proactively in achieving 6-Dof tasks by observing and anticipating their actions without requiring explicit prompts, and humans to be able to provide feedback using physical inputs.  Using a special mount, such tasks includes pickup, dropoff, tilt, and more. 
Compared to language corrections, physical corrections are instantaneous, unambiguous, and robust. This allows the human to effortlessly and precisely correct robot mistakes and guide the robot as needed. However, the challenge no existing methods solve are the following: How to allow an LLM to control the robot while respecting human physical input; How to make the LLM command actions on a semantic level while still enabling rich behaviors in the physical world with centimeter level precision; How to make the human physical correction digestible for the LLM? 

\textbf{Proposed Approach:} Our approach allows a robot to assist humans with a multi-step task, such as cooking, by observing the human and performing reasonable actions as illustrated in Figure \ref{fig: robot_human_cooking}. A LLM, prompted with a semantic description of the scene and the human action, proposes an semantic action translated to an autonomous Dynamical System (DS) action. The robot uses particle filters to model its belief of the DS parameters, and takes the LLM output, treating it as a belief update. If a human physically interacts with the robot, the robot reduces the confidence in the DS parameters and control gains, re-estimating the DS parameters using the velocity as the observation, with minimal human effort. To help LLM in future decision making, a transcript of the interactions along with the corrections is maintained and sent to the LLM when prompting.

Such a system allows a robot to be a helpful assistant and perform collaborative tasks, potentially reducing human effort and task completion time, all without verbal commands. The correction data and the interaction transcript provide more context to the model for subsequent tasks, making the robot more helpful as more tasks are performed. It can also be used to indicate preferences or help avoid mistakes in similar situations. To the best of our knowledge, no previous work has used physical interaction for \textbf{real-time} error correction of LLM-powered robots in a collaborative scenario.

\section{Method}
\begin{figure*}[h]
    \centering
    \includegraphics[width=1.98\columnwidth,trim={0.3cm 0.2cm 0.3cm 0.1cm},clip]{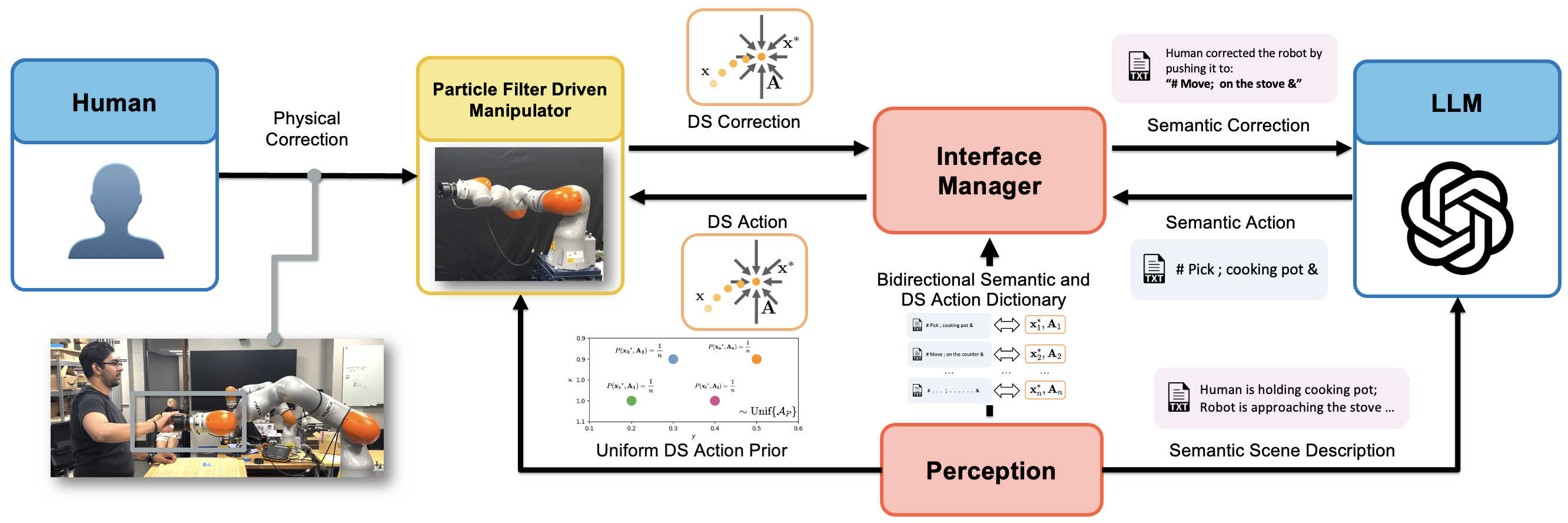}
    \caption{
    \small System overview of our pipeline. The LLM is provided with current semantic scene description from perception module and the previous interaction history. It outputs semantic action that is converted to DS action by the interface manager. This DS action then drives the manipulator by updating the particles. If the human physically corrects the robot, the DS actions are re-estimated based on uniform DS action priors and converted to semantic corrections for the LLM to improve subsequent interactions.}
    \label{fig: system_overview}
\end{figure*}

\subsection{Dynamical System Actions and Semantic Actions}
\label{sec:intent-ds}
\noindent\textbf{Dynamical System Actions}:
     In this work, we assume that setting proper reference velocities for the end effector can help achieve tasks of interest.
     Therefore, we limit the class of reference velocity the end effector tracks to be a \textit{Dynamical System} (DS) \textit{Action} , or requiring the \textbf{end effector} to follow a first-order Cartesian and Rotational autonomous DS 
     \cite{figueroa2020dynamical}, 
\begin{align}
    \dot{\state}\des = \dsf(\state;\dsA,\state^*) = \dsA d(\state, \state^*) \label{eq: ds_model}
\end{align}
    \noindent where $\state = [\xpos, \quat]$ pose state consisting of end effector position $\xpos\in \Rtri$, and orientation $\quat\in \Stri$. $\dot{\state}\des = [\dot{\xpos},\rotv] \in \Rsix $ is the reference linear and rotational velocities given DS parameters $\dsg, \dsA$: $\dsg = [\dsgpos, \dsgquat]$ the position and quaternion \textbf{attractors}, $\dsA\in \Rsixsq$ \textbf{diagonal} negative definite \textbf{dynamics matrices}. $d(\cdot,\cdot)$ is the abstracted difference function as defined in \cite{shao2024constraint} for two pose states.
    While simple, DS has readily available extensions such as \cite{sun2024se, pmlr-v87-figueroa18a} that allow multiple DS to be composed together for more complex non-linear motions, or \cite{li2023task}, which allows editing DS online for obstacle avoidance.

\noindent \textbf{Semantic Action}: Given a list of predefined actions and a list of objects in the scene as perceived by the perception algorithm, we define a semantic action to be performing a valid action on an object, such as ``\textit{Pick up} + \textit{gallon of water}''. The full list of actions, depending on whether the robot is holding an object, include ``\textit{pick}" (to pick up an item), ``\textit{place}" (to put the item down at a specific spot), ``\textit{co-carry}" (to enter co-carry mode for assisting the human with manipulating heavy items), ``\textit{move}" (to move above an item), ``\textit{tilt}" (to tilt by an angle of 20 degrees for pouring),and ``\textit{untilt}" (to revert the tilt). \newline \newline
    The key assumption of our work is the presence of an interface manager as shown in Fig.\ref{fig: system_overview}, which is responsible for converting each DS action to a \textit{semantic action}, and vice versa. 
    We collect perceived object poses and the valid actions on to each object, generating a bidirectional dictionary for corresponding semantic actions and DS actions with parameters $\dsg, \dsA$. The \textit{LLM} is prompted to output one of the valid semantic actions from this dictionary. To convert a semantic action to DS action, the interface manager simply looks up the dictionary and passes the corresponding DS action to the manipulator system. Conversely, for an DS action with parameters $\est\dsg,\est\dsA$, we find the closest DS action $i$ in the dictionary using $\text{argmin}_i(\twonorm{\dsg_i-\est{\dsg}})$, and find the corresponding semantic action.
    
\subsection{Confidence-based Variable Impedance Control}
\label{sec:conf-vic}
During robot operation, the robot tracks a specified DS action and at the same time needs to reduce control effort to allow human physical correction. Therefore, we use confidence-based variable impedance control and adjust the controller gains dynamically. We define confidence measure $c(t)$ as
\begin{align}
    c(t)  = -\int_{-T}^0 \twonorm{\dot \state - \dot \state\des} \dd{s},\label{eq:conf_p}
\end{align}
\noindent where $\dot \state \in \Rsix$ is the linear and rotational velocity of the end effector, $c(t) \in \mathbb R_{\ge 0}$ are clipped to $ [0,1]$, $T$ is the time window of tracking error integration.

Using $c$, we track DS action with the following variable damping only impedance control law inspired by \cite{passiveDS,khoramshahi2020dynamical},
\begin{align}\label{eq:u_ds}
    \dsu &= -c(t) \impD [\dot \state - \dot \state\des]
\end{align}
\noindent where is $\dsu \in \Rsix$ the wrench command in Cartesian space, $\impD \in \Rsixsq$ the constant \textbf{diagonal} negative definite damping gain.
Joint torque $\uj$ can be recovered using $\uj 
 = \jacob(\jntv)^\top \dsu$ with robot Jacobian $\jacob(\jntv)$. More details on how the robot is controlled and the proof of passivity and stability can be found in \cite{shao2024constraint}.

\subsection{Particle Filter Driven Manipulator}\label{subsec: pfga}
To allow response of the LLM response to command the robot, and human input to influence the DS action that is being executed simultaneously, while informed by the available actions in the scene, as shown in Figure \ref{fig: system_overview}, we take a probabilistic approach to estimating DS parameters $\dsg, \dsA$ using a particle filter. On a high level,  the semantic action the LLM outputs is converted to a DS action, and its DS parameters replace all particles' DS parameters. On the other hand, we allow physical corrections to influence the DS parameters during execution of the LLM planned action by using the combination of variable impedance controller and re-estimating the DS parameters with the end-effector velocities in real-time. The following two subsections describes how LLM and human physical interaction influence the DS action parameters.
\subsection{Large Language Model (LLM) as a Collaborative Partner}
We use an LLM to propose reasonable semantic actions given current semantic scene description and interaction history. 

We query the \textit{LLM} when the robot stops at the attractor of a DS action.
In every query, we use a \textbf{system prompt}, shown in Appendix \ref{app: system_prompt}, to guide the \textit{LLM} to have a overview of the task and outputs in a specific format. It specifies the objects in the world and which ones the robot can interact with, It then prioritizes helping the human with heavy or repetitive and is explicitly asked to learn from past corrections. The \textit{LLM} is directed to output a semantic action, along with the reasoning for the decision.

The \textbf{user prompt} consists of the semantic scene description and the history of interaction, and the valid semantic actions in the paired DS action dictionary.
The semantic description of the scene is structured as follows:
1) which objects are currently held by the human and the robot; 
2) the object the human is moving towards, determined using the particle filter method outlined in Sec.\ref{subsec: pfga}; 
3) the robot’s planned action based on previous instructions from the \textit{LLM}; 
4) whether the previous action was successfully executed by robot; 
5) if and where the human physically directs the robot towards a different action physically;
6) all semantic actions available to the robot.

An example of the user prompt is shown here, a sequence of interaction (with interaction history omitted) is presented in Appendix \ref{app: interaction_history}.
\begin{quote}

{\small
\begin{verbatim}
In the previous step, the human did not 
correct the robot’s action.
The last action, "Place on the stove" 
was executed successfully.
The robot is currently holding "Nothing"
The human is holding "Nothing."
The robot is approaching "Nothing."
The human is approaching "Beans"
The available actions are: move "on the 
stove", move "in the sink" ...
\end{verbatim}
}
\end{quote}
As instructed by the system prompt, an example of LLM semantic action response (without reasoning) is

{\small
\begin{verbatim}
# Pick; cooking pot &
\end{verbatim}
} 
We interpret this semantic action by converting it to a DS action by looking it up in the dictionary. All particles states are then replaced with the DS parameters of this DS action, essentially making the robot to executing the LLM proposed semantic action.

\subsection{Human Physical Input as Observations to the Estimation}
When the human is correcting the robot, the action parameter is re-estimated by the particle filter.
We employ a zero dynamics model for the DS parameters, but scale the noise in dynamics based on the confidence. 
The observation is simply the linear and rotational velocity $\dot\state$, with the observation model the same as Eq. \eqref{eq: ds_model}.
The particle weight for particle $i$ is computed as $\exp(-\twonorm{\dot\state - \est{\dot\state}_i}^2)$. To construct the final estimate $\est\dsg, \est\dsA$, we use weighted average of the particles. More details on the particle filter can be found in \cite{shao2024constraint}.

To enhance the convergence of particle filters on meaningful semantic actions when humans interact with the robot, a certain percentage of particles are resampled \textbf{uniformly} on valid DS actions on perceived objects. We set the resample rate $r(t) = 1 - c(t)$, constrained within the interval [0,1], so that the lower the confidence, the more uniformly distributed are the DS parameters estimates.

\begin{figure*}[h]
    \centering
 \includegraphics[width=1.98\columnwidth, trim={0.1cm 0.1cm 0.1cm 0.1cm},clip]{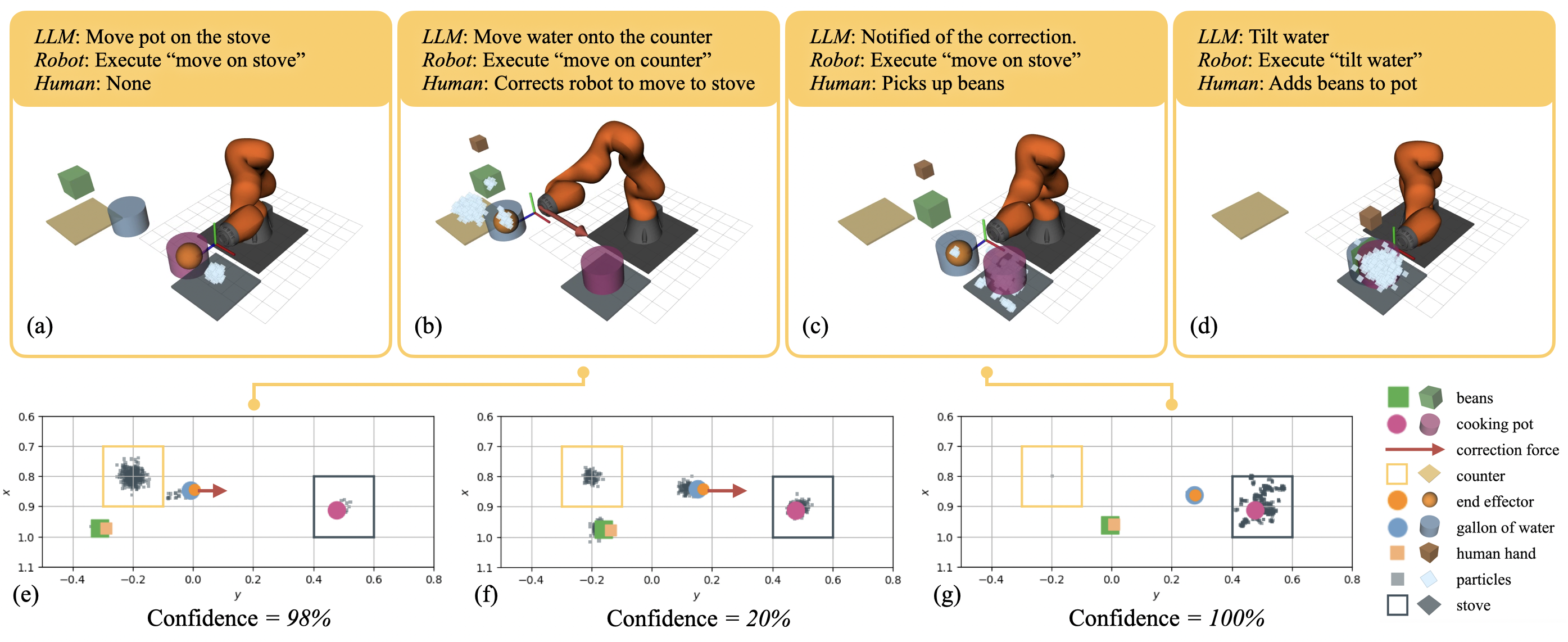}
    \caption{The multi-step task of cooking beans is executed by the LLM-powered robot with physical human correction. Plots \textit{(a),(b),(c),} and \textit{(d)} illustrate the order of task execution with the LLM response, robot action and human activity for key steps. Plots \textit{(e),(f),} and \textit{(g)} show the evolution of DS action parameter belief when physically corrected by the human: The human exert force on the robot, as the tracking error increases, the confidence starts to drop, the robot reduces gains and the particle filter increases resampling rate $r$ by placing particles on perceived objects. Once the human stops interacting with the robot, the tracking error reduces, the confidence and control gains increase. Since most particles are near ``the stove'', the LLM is notified of the correction. } 
    \label{fig:pipeline}
\end{figure*}

A typical correction process pipeline with all components described above is shown in Figure \ref{fig:pipeline}. After correction, when the robot is confident again about the newly estimated DS action, it is converted to a semantic action using the bidirectional DS action dictionary. This semantic action is sent to the LLM as the semantic correction and is saved in the interaction history.

\section{Experiment}
Our aim is to test the ability of the proposed method to proactively assist humans to achieve a multi-step task with physical correction and the interaction history. In the experiments, we want to see 1) if LLM can remember human correction and 2) if the proposed method can smoothly complete multi-step task. 

\subsection{Setup}
We conduct experiments using a hybrid digital twin setup, where a human work alongside a real robot and the objects are virtual. Implementation details can be found in Appendix \ref{app: implement}. We use the Optitrack system to track the human hands for which object the human is apporaching. A 7-\textit{DOF} Kuka iiwa-14 manipulator is controlled using the proposed controller. 
We maintained object positions virtually, updating them if the human or the robot tries to pick one up. For the LLM, we use the GPT-4o\cite{openai2024gpt4} model.

\subsection{Results \& Discussion}
To assess the system's ability to understand human corrections and self-correct for mistakes, using the interaction history, we conduct a text experiment to ascertain if the LLM remembers previous corrections. The result in shown in Table \ref{tab:text_exp}, indicating that corrections in interaction history can help the LLM learn in-context. Appendix \ref{app: interaction_history} shows an example of the LLM's reasoning response.

\begin{table}[h]
\centering
\caption{This table evaluates the LLM's ability to recall previous human corrections after \(n\) steps. The success rate indicates the number of times the LLM suggested the previously human-corrected semantic action when encountering the same semantic state. The results are based on a total of \(20\) trials for each \(n\).}
\label{tab:text_exp}
\begin{tabular}{lcccc}
\toprule
 \textbf{Correction \textit{n} steps ago} & 0 & 5 &  10 & 15 \\
  \textbf{Success Rate} & 100\% & 85\% & 85\% & 80\%  \\
\bottomrule
\end{tabular}
\end{table}

While the LLM suggests reasonable semantic actions most of the time, it can still make mistakes. Without a comprehensive overview of the entire cooking task and only a list of available semantic actions, especially at the beginning of a multi-step task, it may generate infeasible or unreasonable actions such as getting stuck in a loop of two semantic actions. We assess the ability of whether physical human correction allow intuitive change in DS actions and as a result makes the robot a helpful assistant. We observe that the robot adapts to the new action with minimal human effort and notice that the robot's belief is sensitive to the prior distribution over objects in the scene, as illustrated in Figure \ref{fig:pipeline}. This highlights the proposed method's capability to adapt dynamically, ensuring alignment with human preferences and enhancing overall system usefulness.

\section{Future Works}
While the current work has demonstrated the exciting potential of LLM-powered robots to physically interact with humans, there are many details left to be explored. These include designing robust grasping hardware and algorithms to allow fast DS action for pickup. Additionally, there are alternative methods for integrating LLM corrections. For example, instead of recording the semantic correction as part of the interaction history, it is also possible to use the correction as part of the fine-tuning or RAG data. 


\bibliographystyle{unsrt}
\bibliography{references}

\begin{thebibliography}{10}

\bibitem{octo_2023}
{Octo Model Team}, Dibya Ghosh, Homer Walke, Karl Pertsch, Kevin Black, Oier Mees, Sudeep Dasari, Joey Hejna, Charles Xu, Jianlan Luo, Tobias Kreiman, {You Liang} Tan, Lawrence~Yunliang Chen, Pannag Sanketi, Quan Vuong, Ted Xiao, Dorsa Sadigh, Chelsea Finn, and Sergey Levine.
\newblock Octo: An open-source generalist robot policy.
\newblock In {\em Proceedings of Robotics: Science and Systems}, Delft, Netherlands, 2024.

\bibitem{ahn2022i}
Michael Ahn, Anthony Brohan, Noah Brown, Yevgen Chebotar, Omar Cortes, Byron David, Chelsea Finn, Chuyuan Fu, Keerthana Gopalakrishnan, Karol Hausman, Alex Herzog, Daniel Ho, Jasmine Hsu, Julian Ibarz, Brian Ichter, Alex Irpan, Eric Jang, Rosario~Jauregui Ruano, Kyle Jeffrey, Sally Jesmonth, Nikhil~J Joshi, Ryan Julian, Dmitry Kalashnikov, Yuheng Kuang, Kuang-Huei Lee, Sergey Levine, Yao Lu, Linda Luu, Carolina Parada, Peter Pastor, Jornell Quiambao, Kanishka Rao, Jarek Rettinghouse, Diego Reyes, Pierre Sermanet, Nicolas Sievers, Clayton Tan, Alexander Toshev, Vincent Vanhoucke, Fei Xia, Ted Xiao, Peng Xu, Sichun Xu, Mengyuan Yan, and Andy Zeng.
\newblock Do as i can, not as i say: Grounding language in robotic affordances, 2022.

\bibitem{huang2022inner}
Wenlong Huang, Fei Xia, Ted Xiao, Harris Chan, Jacky Liang, Pete Florence, Andy Zeng, Jonathan Tompson, Igor Mordatch, Yevgen Chebotar, Pierre Sermanet, Noah Brown, Tomas Jackson, Linda Luu, Sergey Levine, Karol Hausman, and Brian Ichter.
\newblock Inner monologue: Embodied reasoning through planning with language models, 2022.

\bibitem{Wu_2023}
Jimmy Wu, Rika Antonova, Adam Kan, Marion Lepert, Andy Zeng, Shuran Song, Jeannette Bohg, Szymon Rusinkiewicz, and Thomas Funkhouser.
\newblock Tidybot: personalized robot assistance with large language models.
\newblock {\em Autonomous Robots}, 47(8):1087–1102, November 2023.

\bibitem{ma2024eureka}
Yecheng~Jason Ma, William Liang, Guanzhi Wang, De-An Huang, Osbert Bastani, Dinesh Jayaraman, Yuke Zhu, Linxi Fan, and Anima Anandkumar.
\newblock Eureka: Human-level reward design via coding large language models, 2024.

\bibitem{liang2023code}
Jacky Liang, Wenlong Huang, Fei Xia, Peng Xu, Karol Hausman, Brian Ichter, Pete Florence, and Andy Zeng.
\newblock Code as policies: Language model programs for embodied control, 2023.

\bibitem{embodimentcollaboration2024open}
Embodiment Collaboration.
\newblock Open x-embodiment: Robotic learning datasets and rt-x models, 2024.

\bibitem{lynch2022interactive}
Corey Lynch, Ayzaan Wahid, Jonathan Tompson, Tianli Ding, James Betker, Robert Baruch, Travis Armstrong, and Pete Florence.
\newblock Interactive language: Talking to robots in real time, 2022.

\bibitem{Cui_2023}
Yuchen Cui, Siddharth Karamcheti, Raj Palleti, Nidhya Shivakumar, Percy Liang, and Dorsa Sadigh.
\newblock No, to the right: Online language corrections for robotic manipulation via shared autonomy.
\newblock In {\em Proceedings of the 2023 ACM/IEEE International Conference on Human-Robot Interaction}, HRI ’23. ACM, March 2023.

\bibitem{shi2024yell}
Lucy~Xiaoyang Shi, Zheyuan Hu, Tony~Z. Zhao, Archit Sharma, Karl Pertsch, Jianlan Luo, Sergey Levine, and Chelsea Finn.
\newblock Yell at your robot: Improving on-the-fly from language corrections, 2024.

\bibitem{liu2023reflect}
Zeyi Liu, Arpit Bahety, and Shuran Song.
\newblock Reflect: Summarizing robot experiences for failure explanation and correction, 2023.

\bibitem{zha2024distilling}
Lihan Zha, Yuchen Cui, Li-Heng Lin, Minae Kwon, Montserrat~Gonzalez Arenas, Andy Zeng, Fei Xia, and Dorsa Sadigh.
\newblock Distilling and retrieving generalizable knowledge for robot manipulation via language corrections, 2024.

\bibitem{zheng2024evaluating}
Zhi Zheng, Qian Feng, Hang Li, Alois Knoll, and Jianxiang Feng.
\newblock Evaluating uncertainty-based failure detection for closed-loop llm planners, 2024.

\bibitem{jiang2024transic}
Yunfan Jiang, Chen Wang, Ruohan Zhang, Jiajun Wu, and Li~Fei-Fei.
\newblock Transic: Sim-to-real policy transfer by learning from online correction, 2024.

\bibitem{losey2022physical}
Dylan~P. Losey, Andrea Bajcsy, Marcia~K. O’Malley, and Anca~D. Dragan.
\newblock Physical interaction as communication: Learning robot objectives online from human corrections.
\newblock {\em The International Journal of Robotics Research}, 41(1):20--44, 2022.

\bibitem{pmlr-v48-finn16}
Chelsea Finn, Sergey Levine, and Pieter Abbeel.
\newblock Guided cost learning: Deep inverse optimal control via policy optimization.
\newblock In Maria~Florina Balcan and Kilian~Q. Weinberger, editors, {\em Proceedings of The 33rd International Conference on Machine Learning}, volume~48 of {\em Proceedings of Machine Learning Research}, pages 49--58, New York, New York, USA, 20--22 Jun 2016. PMLR.

\bibitem{bobu2021feature}
Andreea Bobu, Marius Wiggert, Claire Tomlin, and Anca~D Dragan.
\newblock Feature expansive reward learning: Rethinking human input.
\newblock In {\em Proceedings of the 2021 ACM/IEEE International Conference on Human-Robot Interaction}, pages 216--224, 2021.

\bibitem{li2021learning}
Mengxi Li, Alper Canberk, Dylan~P Losey, and Dorsa Sadigh.
\newblock Learning human objectives from sequences of physical corrections.
\newblock In {\em 2021 IEEE International Conference on Robotics and Automation (ICRA)}, pages 2877--2883. IEEE, 2021.

\bibitem{figueroa2020dynamical}
Nadia Figueroa, Salman Faraji, Mikhail Koptev, and Aude Billard.
\newblock A dynamical system approach for adaptive grasping, navigation and co-manipulation with humanoid robots.
\newblock In {\em 2020 IEEE International conference on robotics and automation (ICRA)}, pages 7676--7682. IEEE, 2020.

\bibitem{shao2024constraint}
Yifei Simon~Shao, Tianyu Li, Shafagh Keyvanian, Pratik Chaudhari, Vijay Kumar, and Nadia Figueroa.
\newblock Constraint-aware intent estimation for dynamic human-robot object co-manipulation.
\newblock In {\em Robotics: Science and Systems}, 2024.

\bibitem{sun2024se}
Sunan Sun and Nadia Figueroa.
\newblock Se (3) linear parameter varying dynamical systems for globally asymptotically stable end-effector control.
\newblock {\em arXiv preprint arXiv:2403.16366}, 2024.

\bibitem{pmlr-v87-figueroa18a}
Nadia Figueroa and Aude Billard.
\newblock A physically-consistent bayesian non-parametric mixture model for dynamical system learning.
\newblock In Aude Billard, Anca Dragan, Jan Peters, and Jun Morimoto, editors, {\em Proceedings of The 2nd Conference on Robot Learning}, volume~87 of {\em Proceedings of Machine Learning Research}, pages 927--946. PMLR, 29--31 Oct 2018.

\bibitem{li2023task}
Tianyu Li and Nadia Figueroa.
\newblock Task generalization with stability guarantees via elastic dynamical system motion policies.
\newblock In {\em 7th Annual Conference on Robot Learning}, 2023.

\bibitem{passiveDS}
Klas Kronander and Aude Billard.
\newblock Passive interaction control with dynamical systems.
\newblock {\em IEEE Robotics and Automation Letters}, 1(1):106--113, 2016.

\bibitem{khoramshahi2020dynamical}
Mahdi Khoramshahi and Aude Billard.
\newblock A dynamical system approach for detection and reaction to human guidance in physical human--robot interaction.
\newblock {\em Autonomous Robots}, 44(8):1411--1429, 2020.

\bibitem{openai2024gpt4}
OpenAI.
\newblock Gpt-4 technical report, 2024.

\end{thebibliography}

\appendix

  

\subsection{LLM System Prompt}
\label{app: system_prompt}

We use the following system prompt to encourage the model to aid a human in the kitchen environment and learn from human corrections. The system prompt is presented below: 

\begin{quote}
\textbf{Role:} You are a robotic assistant named ChefBot, tasked with aiding a human in the kitchen environment.

\textbf{Objective:} Your mission is to facilitate kitchen tasks effectively, focusing on optimal interaction with items and the environment.

\textbf{Item Categories:}

\begin{itemize}
    \item \textbf{Category A (Items with Mount):} These are items the robot can pick up, typically containers. (Examples: 'gallon of water', 'cooking pot', 'cutting board')
    \item \textbf{Category B (Environment Items):} Places where items can be set down when held. (Examples: 'on the stove', 'in the sink', 'on the counter')
    \item \textbf{Category C (Items without Mount):} Food items that can only be manipulated when placed atop a Category A item. (Examples: 'lettuce', 'chicken breast', 'beans')
\end{itemize}

\textbf{Abilities:}

\begin{itemize}
    \item \textbf{Pick:} Executable only when the robot is empty-handed and over a Category A item, combining Move and Pick actions.
    \item \textbf{Move:} Allows navigation over any item.
    \item \textbf{Place:} Places items in hand at a Category B location. Placement should be generic, not specific.
    \item \textbf{Tilt/Untilt:} Enables tilting objects held over a Category A item and subsequently reverting them to their original state.
\end{itemize}

\textbf{Operation Instructions:}

\begin{itemize}
    \item \textbf{Feedback Learning:} Absorb lessons from human corrections and action feedback to refine actions independently of direct interventions.
    \item \textbf{Action Execution:} Always begin actions with '\#', separate commands and items with ';', and conclude with '\&'.
    \item \textbf{Response Requirement:} Every action response must include a reasoning step, clarifying the robot's decision-making process.
\end{itemize}

\textbf{Example Command:} \# Pick ; cooking pot \&

\textbf{Special Notes:}

\begin{itemize}
    \item One Command at a time.
    \item Human corrections: Human will directly correct you to the right place. If last time human corrected you on the stove, next time when the same state happens, considering the command "go to the stove".
    \item Interaction history and feedback on action results are provided. Use this information to improve performance.
    \item Human Priority: Always prioritize assisting the human collaboratively and efficiently.
\end{itemize}
\end{quote}

\subsection{Interaction history}
\label{app: interaction_history}

This is an example of interaction history in our experiment.

\subsubsection*{Interaction 3}\textbf{Prompt:}

\begin{quote}
In the previous step, the human did not correct the robot's action. The robot is holding 'cooking pot'. The human is holding 'Nothing'. The robot is approaching 'Nothing'. The human is approaching 'Nothing'. The available actions are: place 'on the stove', place 'on the counter', tilt 'cooking pot', untilt 'cooking pot', circular 'cooking pot', move 'gallon of water', move 'cooking pot', move 'on the stove', move 'on the counter'.
\end{quote}

\textbf{Suggested Robot Action:}
\begin{verbatim}
# Place ; on the counter &
\end{verbatim}

\textbf{Reasoning:}
\begin{enumerate}
    \item The robot is holding the cooking pot and the human is not holding anything.
    \item Placing the cooking pot on the counter will make it easily accessible for the human to place the 'beans' into it or to perform other tasks related to cooking.
    \item This action allows for flexibility in case the human needs to use the cooking pot immediately.
\end{enumerate}

\subsubsection*{Interaction 4}
\textbf{Prompt:}
\begin{quote}
In the previous step, the human corrected the robot's action by pushing it to: 'on the stove'. The final action executed by the robot was: Move 'on the stove'. The robot is holding 'cooking pot'. The human is holding 'Nothing'. The robot is approaching 'on the counter'. The human is approaching 'Nothing'. The available actions are: place 'on the stove', place 'on the counter', tilt 'cooking pot', untilt 'cooking pot', circular 'cooking pot', move 'gallon of water', move 'cooking pot', move 'on the stove', move 'on the counter'.
\end{quote}

\textbf{Suggested Robot Action:}
\begin{verbatim}
# Place ; on the stove &
\end{verbatim}

\textbf{Reasoning:}
\begin{enumerate}
    \item The human corrected the robot's previous action to move to the stove.
    \item When last corrected, the human wanted the cooking pot to be placed on the stove.
    \item Therefore, placing the cooking pot on the stove is the optimal action to align with the human's preferences.
\end{enumerate}

\subsection{Implementation}
\label{app: implement}

We borrow the implementation from \cite{shao2024hriworkshop} for the proposed method. The robot controller (200 Hz) (DS Variable Impedance Controller) is implemented in C++ using the open-source ROS stack \textit{iiwa\_ros}\footnote{\url{https://github.com/epfl-lasa/iiwa_ros}}. The particle filters (20 Hz) are implemented in Python with significant changes on top of \textit{pfilter}\footnote{\url{https://pypi.org/project/pfilter/}}. 
  Experiments are performed on a desktop with Nvidia GeForce RTX-3070. The proposed method hyperparameters are reported in Table \ref{tb: var_gain}. The simulation environment is setup in \textit{gazebo} with a 7-\textit{DOF} Kuka iiwa-14 manipulator and the human force on the end-effector is passed in with a joystick.
\begin{table}[h]
\centering
\begin{tabular}{|l|l|l|}
\hline
                       & Low  & High   \\ \hline
Damping Gain Cartesian $\impD_\pos$ & 1    & 85     \\ \hline
Damping Gain Rotation $\impD_\rot$  & 1    & 13     \\ \hline
Dynamics Cartesian $\dsAp$    & -0.6 & -0.4   \\ \hline
Dynamics Rotation  $\dsAr$     & -0.9 & -0.6   \\ \hline
Relative Weights in Particle Filter $\eta_1 \sim \eta_4$  & 0.5    &      \\ \hline
Ellipsoid Decay Weight $\tune_5$       & 1.5 &    \\ \hline
Noise Dynamics Cartesian $\tune_6$         & 3e-4 & 4e-3 \\ \hline
Noise Dynamics Rotation $\tune_7$         & 2e-4 & 8.5e-3 \\ \hline
Noise Goal Rotation $\tune_8$         & 2e-4 & 8.5e-3 \\ \hline
Joint Limit Resistance $\eta_9$         & 0.1& \\ \hline
Ascent Rate Cartesian $\decay_p$ &  0.41 &\\ \hline
Ascent Rate Rotation $\decay_o$ & 0.49 &\\
\hline
\end{tabular}
\caption{Hyperparameters. Fixed values are written in low column}
\label{tb: var_gain}
\end{table}

\end{document}